\documentclass[11pt]{article}
\usepackage{mtsummit2019}
\usepackage{times}
\usepackage{url}
\usepackage{latexsym}
\usepackage[small,bf]{caption} 
\usepackage[utf8]{inputenc}
\usepackage[T1]{fontenc}
\usepackage{multirow}
\usepackage{booktabs}
\usepackage{subcaption}
\usepackage{suffix}
\usepackage{xcolor}
\usepackage{natbib}

\newcommand{\Table}[1]{Table~\ref{tab:#1}}

\newcommand{\ie}{i.\,e.,\ }
\WithSuffix\newcommand\ie*{i.\,e.,}
\newcommand{\Ie}{I.\,e.,\ }
\WithSuffix\newcommand\Ie*{I.\,e.,}
\newcommand{\eg}{e.\,g.,\ }
\WithSuffix\newcommand\eg*{e.\,g.,}
\newcommand{\Eg}{E.\,g.,\ }
\WithSuffix\newcommand\Eg*{E.\,g.,}
\newcommand{\ibid}{ibid.\ }
\WithSuffix\newcommand\ibid*{ibid.}
\newcommand{\Ibid}{Ibid.\ }
\WithSuffix\newcommand\Ibid*{Ibid.}
\newcommand{\percent}{\,\%\ }
\WithSuffix\newcommand\percent*{\,\%}
\newcommand{\etal}{et al.\ }
\WithSuffix\newcommand\etal*{et al.}

\newcommand{\tm}{\textsc{TM-Only} }
\WithSuffix\newcommand\tm*{\textsc{TM-Only}}
\newcommand{\pe}{\textsc{Post-Edit} }
\WithSuffix\newcommand\pe*{\textsc{Post-Edit}}

\newcommand{\thecompany}{Migros Bank }
\WithSuffix\newcommand\thecompany*{Migros Bank}

\title{Post-editing Productivity with Neural Machine Translation: An Empirical Assessment of Speed and Quality in the Banking and Finance Domain}

\author{}

\author{Samuel Läubli\,$^{1,3}$ 
    \quad Chantal Amrhein\,$^{1,3}$
    \quad Patrick Düggelin\,$^1$\medskip\\
    \quad \textbf{Beatriz Gonzalez\,$^2$}
    \quad \textbf{Alena Zwahlen\,$^1$}
    \quad \textbf{Martin Volk\,$^{1,3}$} \bigskip\\
    $^1$\,TextShuttle AG \quad
    $^2$\,Migros Bank AG\medskip\\
    $^3$\,Institute of Computational Linguistics, University of Zurich}

\date{}

\begin{document}
\maketitle
\begin{abstract}
  Neural machine translation (NMT) has set new quality standards in automatic translation, yet its effect on post-editing productivity is still pending thorough investigation. We empirically test how the inclusion of NMT, in addition to domain-specific translation memories and termbases, impacts speed and quality in professional translation of financial texts. We find that even with language pairs that have received little attention in research settings and small amounts of in-domain data for system adaptation, NMT post-editing allows for substantial time savings and leads to equal or slightly better quality.
\end{abstract}

\section{Introduction}
\label{sec:Introduction}

The use of neural networks for sequence transduction \citep{KalchbrennerBlunsom2013,Sutskever2014,Bahdanau2015} has led to astounding progress in the field of machine translation (MT), establishing a new level of quality in applications such as automatic news translation \citep{Sennrich2016,Hassan2018}. Nevertheless, the creation of publication-grade translations still requires human involvement \citep{Laeubli2018,Toral2018}, and previous work has explored human--machine collaboration in the form of post-editing, where human experts revise machine translated text where necessary.

Empirical investigations of post-editing productivity with NMT are still scarce, especially for language combinations that do not include English as either the source or target language. In this paper, we describe and discuss the results of a productivity test of domain-adapted NMT with the in-house translation team of Migros Bank, a mid-sized financial institution based in Zurich, Switzerland. We evaluate the use of NMT under typical working conditions, focussing on two rarely explored language pairs---German~(DE) to French~(FR) and Italian~(IT)---and texts from a specialised domain: banking and finance. We show that using NMT in combination with translation memories (TMs) and terminology databases (TBs) enables professional translators to work faster with no loss (or slight gains) in quality, even with limited in-domain data for system adaptation.



\begin{table*}[t]
    \centering
    \renewcommand{\arraystretch}{1.2}
    \fontsize{10.1pt}{10.1pt}\selectfont
    \begin{tabular}{lll rrrr c rrrr}
    \toprule
    Text    & Topic     & Source Words  & \multicolumn{4}{c}{Coverage FR}                   && \multicolumn{4}{c}{Coverage IT}                       \\[0.5mm]
            &           &               & \multicolumn{1}{c}{100\percent} 
                                                      & \multicolumn{1}{c}{90\percent}
                                                                   & \multicolumn{1}{c}{80\percent} 
                                                                                & \multicolumn{1}{c}{$R$}       
                                                                                            && \multicolumn{1}{c}{100\percent} 
                                                                                                             & \multicolumn{1}{c}{90\percent}
                                                                                                                             & \multicolumn{1}{c}{80\percent}
                                                                                                                                             & \multicolumn{1}{c}{$R$} \\
    \midrule
    A       & Funding   & 1108          & 8.94        & 0.54       & 6.59       & 14.69     && 9.57          & 1.17          & 6.59          & 15.89 \\
    B       & Funding   & 1006          & 2.58        & 3.68       & 4.17       & 9.23      && 1.29          & 1.69          & 1.39          & 3.93  \\
    C       & Investing & 1059          & 6.80        & 2.08       & 1.89       & 10.18     && 4.25          & 3.31          & 3.02          & 9.64  \\
    D       & Investing & 1077          & 14.48       & 10.58      & 0.84       & 24.68     && 7.24          & 15.51         & 3.06          & 23.65 \\
    \bottomrule
\end{tabular}
    \caption{Source texts (DE) and their TM coverage in the target languages (FR, IT). Fuzzy bands include 90.00--99.99\percent (referred to as 90\percent*) and 80.00--89.99\percent (referred to as 80\percent*) matches. Coverage coefficients $R$ indicate the percentage of translated words available in the TM for each text, considering 80--100\percent matches.}
    \label{tab:Texts}
\end{table*}

\section{Background}
\label{sec:Background}

Early assessments of post-editing productivity were focussed on technical texts. While a study by \citet{Krings1994} with user manuals for technical appliances and rule-based MT found mixed results, interest in post-editing grew with the advent of statistical MT (SMT), which enabled time savings of up to 40\percent in film subtitling \citep{Volk2008,Sousa2011} and software localisation \citep{PlittMasselot2010}. Subsequent work concluded that significant time savings can also be achieved in more complex domains such as legal \citep{Federico2012} or marketing texts \citep{Laeubli2013}.

Many productivity tests explored either translation from or into English \citep[\eg][]{PlittMasselot2010}, or translation between closely related languages such as Swedish and Danish \citep[\eg][]{Volk2008}. \citet{Green2013} conducted a large-scale experiment from English into three target languages with different canonical word order: Arabic (VSO), French (SVO), and German (SOV). While post-editing was significantly faster than translation from scratch for all combinations, it is unclear whether their findings would equally apply to language pairs that do not include English, particularly if less MT training material is available. We investigate two language pairs that have received little attention in post-editing research: DE--FR and DE--IT.

The effect of using NMT rather than SMT on translation productivity has not yet been conclusively assessed. One of the first studies contrasting NMT and SMT quality found that NMT produces less morphological, lexical, and word order errors, thus reducing post-editing effort by 26\percent in English to German subtitle translation \citep{Bentivogli2016}. However, post-editing effort was measured with HTER \citep{Snover2006}, a distance-based metric. \citet{Castilho2017} found that although more fluent, post-editing NMT rather than SMT output did not save time in an educational domain due to a higher number of omissions, additions, and mistranslations. Conversely, time savings doubled with NMT (+36\percent*) compared to SMT (+18\percent*) in literary translation \citep{Toral2018Literary}. The number of studies on NMT post-editing productivity is still limited, and further studies are needed, not least because findings obtained with different domains and language combinations are difficult to compare. The present study contributes data on NMT post-editing speed and quality in the financial domain.

Previous productivity tests used different experimental designs. In early work, \citet{Krings1994} found that post-editing of rule-based MT resulted in a decrease in translation time by 7\percent when translators used pen and paper, but an increase by 20\percent when they used a computer instead. \citet{PlittMasselot2010} and \citet{Green2013} compared post-editing to translation from scratch, using purpose-built web interfaces that showed one source sentence at a time, paired with a target text box that was either populated with MT or empty. Proponents of field tests have argued that while improving control of extraneous variables, such designs reduce experimental validity in that they isolate translators from tools long indispensable in professional workflows, namely software workbenches that show multiple sentences at a time and suggestions from TMs and TBs \citep{Federico2012,Laeubli2013}. We chose an in-situ design where translators had access to the tools and resources known from their daily work.

\begin{table*}[t]
    \centering
    \begin{subtable}[b]{0.48\textwidth}
        \centering
        \renewcommand{\arraystretch}{1.2}
    \fontsize{10.1pt}{10.1pt}\selectfont
        \begin{tabular}{llllrr}
\toprule
Subject    & Text    & Seq.\    & MT  & Words/h & Quality \\
\midrule
FR-1       & A       & 1        & No  & 520.37  & 4.00    \\
FR-1       & B       & 2        & No  & 630.82  & 5.50    \\
FR-1       & C       & 3        & Yes & 909.88  & 5.00    \\
FR-1       & D       & 4        & Yes & 602.56  & 5.00    \\
FR-2       & A       & 1        & Yes & 987.00  & 4.50    \\
FR-2       & B       & 2        & Yes & 1237.13 & 3.50    \\
FR-2       & C       & 3        & No  & 682.64  & 4.00    \\
FR-2       & D       & 4        & No  & 505.40  & 4.50    \\
\midrule
\multicolumn{3}{l}{Average \tm} & No  & 584.81  & 4.50    \\
\multicolumn{3}{l}{Average \pe} & Yes & 934.14  & 4.50    \\
\multicolumn{4}{l}{Difference (\%)}   & 59.74   & 0.00    \\
\bottomrule
\end{tabular}
        \caption{DE--FR}
        \label{tab:ResultsFR}
    \end{subtable}
    \quad
    \begin{subtable}[b]{0.48\textwidth}
        \centering
        \renewcommand{\arraystretch}{1.2}
        \fontsize{10.1pt}{10.1pt}\selectfont
        \begin{tabular}{llllrr}
\toprule
Subject    & Text    & Seq.\    & MT  & Words/h & Quality \\
\midrule
IT-1       & A       & 1        & No  & 389.41  & 4.00    \\
IT-1       & B       & 2        & No  & 398.71  & 4.00    \\
IT-1       & C       & 3        & Yes & 647.87  & 4.50    \\
IT-1       & D       & 4        & Yes & 393.14  & 4.00    \\
IT-2       & A       & 1        & Yes & 401.19  & 5.50    \\
IT-2       & B       & 2        & Yes & 536.09  & 5.50    \\
IT-2       & C       & 3        & No  & 553.00  & 5.50    \\
IT-2       & D       & 4        & No  & 469.56  & 5.50    \\
\midrule
\multicolumn{3}{l}{Average \tm} & No  & 452.67  & 4.75    \\
\multicolumn{3}{l}{Average \pe} & Yes & 494.57  & 4.88    \\
\multicolumn{4}{l}{Difference (\%)}   & 9.26    & 0.13    \\
\bottomrule
\end{tabular}
        \caption{DE--IT}
        \label{tab:ResultsIT}
    \end{subtable}
    \caption{Experimental conditions and results: the number of target words produced per hour (Words/h) and averaged overall impression scores (Quality) as assigned by two expert raters per translation.}
    \label{tab:Results}
\end{table*}

\section{Assessment of Translation Productivity}
\label{sec:Main}

We conducted a productivity test of domain-adapted NMT on the premises of \thecompany*. Subjects translated texts under two experimental conditions. In \tm*, they used the translation workbench known from their their daily work, including a domain-specific TM, a domain-specific TB, and any online services (except machine translation) of choice. The same setup was used in \pe*, except that sentences with no fuzzy match of at least 80\percent in the TM were populated with MT within the translation workbench. We did not show MT where high fuzzy matches were available because editing high fuzzy matches is more efficient \citep{SanchezGijon2019}.

\paragraph{Materials} We used four German source texts from \thecompany*. The texts had not been translated by any of the translators involved in the experiment before, and had been excluded from the MT training material (see below). The TMs contained several exact and high fuzzy matches for each text (\Table{Texts}).

To pretranslate sentences in \pe*, we trained WMT17-style bi-RNN systems \citep{Sennrich2017wmt} using the \texttt{marian} toolkit \citep{Junczys2016}. The training material consisted of 6 million out-of-domain segments from publicly available OPUS corpora \citep{OPUS}, as well as 385'320 and 186'647 in-domain segments for FR and IT, respectively. We filtered both in- and out-of-domain segments through a set of mostly length-based heuristics \citep{zwahlen2016tm}, and oversampled the former as a simple means of domain adaptation. While this has proven effective in other contexts \citep[\eg*][]{Sennrich2016WMT}, we note that translation quality could likely be improved by means of more advanced techniques such as fine-tuning \citep{LuongManning2015} or multi-domain modelling \citep{Chu2017}.

\paragraph{Subjects} A total of four professional translators took part in the productivity test, two each for the target languages FR (FR-1, FR-2) and IT (IT-1, IT-2). All were members of \thecompany*'s internal translation team. They were therefore familiar both with the software used and with the language and terminology of the documents to be translated. FR-1, who joined the organisation shortly before the experiment, was less experienced than the other participants. All subjects had been post-editing outputs of the MT systems used in the experiment (see above) for three months, and had received four hours of post-editing training.

\paragraph{Procedure}

Each subject translated the four German source texts in the same order. Conditions were counterbalanced (\Table{Results}). Subjects were first briefed about the purpose and data collected during the experiment. They were then given 60 minutes to work on each text, which we announced would likely not be enough to translate all sentences. There were 10-minute breaks between working blocks, and a 30-minute break in the middle of the experiment. A post-experimental survey concluded the experiment.

We encountered no problems with data collection, with the exception of a temporary failure of IT-1's screen in the last working block. The device went into standby mode, which was not reported immediately and resulted in a total interruption of 4 minutes, which we deducted from the respective session before calculating translation speed as shown in \Table{Results}.


\begin{table*}[t]
    \centering
    \renewcommand{\arraystretch}{1.2}
    \fontsize{10.1pt}{10.1pt}\selectfont
    \begin{tabular}{lcccc}
    \toprule
    Criterion          & \multicolumn{2}{c}{DE--FR} & \multicolumn{2}{c}{DE--IT} \\[0.5mm]
                       & \tm        & \pe       & \tm        & \pe       \\
    \midrule
    Coherence          & 4.75       & 5.25      & 5.00       & 5.00      \\
    Cohesion           & 4.75       & 4.50      & 5.25       & 5.00      \\
    Grammar            & 4.75       & 4.75      & 4.75       & 4.88      \\
    Style              & 4.50       & 5.00      & 5.00       & 5.00      \\
    Cultural adequacy  & 4.50       & 4.75      & 4.50       & 4.75      \\
    \midrule
    Overall Impression & 4.50       & 4.50      & 4.75       & 4.88      \\
    \bottomrule
\end{tabular}
    \caption{Detailed quality assessment results. Each cell is an average over eight scores: four translations scored by two expert raters. Overall impression was graded separately; it is not an average over the other criteria.}
    \label{tab:Quality}
\end{table*}

\subsection{Speed}
\label{sec:Speed}

We report translation speed as the number of target words produced per hour. To account for TM matches, we derive a TM coverage coefficient $R$ for each text:

\begin{equation}
    R = 1a + 0.9b + 0.8c,
\end{equation}

\noindent where $a$ is the percentage of 100\percent*, $b$ the percentage of 90\percent*, and $c$ the percentage of 80\percent TM matches. We then adjust the number of words $W$ translated in each experimental block as

\begin{equation}
    W^{*} = (1-R)~W.
\end{equation}

\noindent This approximation assumes uniform distribution of TM matches within texts.

Results are shown in \Table{Results}. FR subjects produced 584.81 and 934.14 words per hour in \tm and \pe*, respectively, an increase of 59.74\percent*. The difference was less marked in IT, with 452.67 and 494.57 words per hour produced in \tm and \pe*, respectively (9.26\percent*).

While focussing on descriptive statistics due to small sample size, we also fit linear-mixed effects models for inferential analysis. \citet{CarterWojton2018} show that very small sample sizes can attain sufficient power when a single fixed effect factor is of interest, albeit at a greater risk of type I errors. We use experimental condition (\tm vs.\ \pe*) as the fixed effect factor, and random intercepts for subjects and texts. The models show no deviation from homoscedasticity or normality in visual inspection of residual plots and Shapiro-Wilk tests. Likelihood ratio tests show a significant main effect of experimental condition in FR ($\chi^2(1)=9.74, p<.01$), but not IT ($\chi^2(1)=0.93, p=.33$).

\subsection{Quality}
\label{sec:Quality}

The translations produced in the experiment were reviewed by university lecturers in professional translation, who were remunerated at standard hourly rates. Experts did not know which translations were produced using MT. The quality of each translation was independently assessed by two experts, who assigned scores on a 6-point scale (1 = worst, 6 = best) for coherence, cohesion, grammar, style, cultural adequacy, and overall impression. 

Results are shown in \Table{Quality}. Each cell is an average over 8 scores: 4 texts evaluated by two experts. Note that experts assigned separate scores for overall impression, which may therefore deviate from the average over scores for the other criteria. Average per-text scores for overall impression are included in \Table{Results}.

Considering overall impression, experts did not find a difference in quality between texts produced with and without MT in FR. In IT, texts translated with MT received slightly higher scores ($+0.13$). MT improved coherence ($+0.50$), style ($+0.50$), and cultural adequacy ($+0.25$) in FR, as well as grammar ($+0.13$) and cultural adequacy ($+0.25$) in IT. Cohesion, on the other hand, was found to be better in texts produced without MT in both FR ($-0.25$) and IT ($-0.25$).

\section{Discussion}
\label{sec:Discussion}

While the minimum speed hardly differed between \tm and \pe, the latter allowed for higher average and maximum speed. In FR, the highest speed measured in \pe was 1237.13 words per hour (FR-2, text B), as opposed to 682.64 words per hour in \tm (FR-2, text C). In IT, the maximum speed in \pe was 647.87 words (IT-1, text C), and 553.00 words per hour in \tm (IT-2, text C).

Three out of four translators were faster in \pe on average. IT-2 did not benefit from MT: With an average speed of 511.28 words per hour in \tm and 468.64 words per hour in \pe*, the subject was 8.34\percent slower. Previous research has shown that not all translators benefit equally from MT \citep[\eg][]{PlittMasselot2010,KoehnGermann2014}, which calls for large sample sizes in productivity tests \citep{Green2013}. Although improving robustness, involving a large number of translators is not always possible in practice -- in our case, the in-house translation team had no more than four members, and involving external translators would have introduced other confounds (such as domain knowledge) that are hard to control for. We also note that IT-2 produced translations of above-average quality (Tables~\ref{tab:Results},~\ref{tab:Quality}), suggesting that MT may be less beneficial when aiming for maximum quality.

Another observation that warrants discussion is the difference in productivity between the two target languages. Again, one possible explanation is the small number of participants and measurements. A larger number of measurements would allow more accurate conclusions to be drawn as to whether the maximum speed achieved in FR (FR-2, text B) is to be treated as an outlier, or if translators will repeatedly achieve a throughput of more than 1,000 words per hour with MT. Moreover, the DE--IT engine was trained with less in-domain material than the DE--FR engine. This resulted in lower raw MT quality for IT, which in turn may have resulted in lower productivity.\footnote{However, \citet{KoehnGermann2014} find that between-subjects variance is higher than between-systems variance in post-editing.} Screen recordings also showed that IT translators made more stylistic changes to MT outputs, but apart from slightly higher quality scores overall (\Table{Quality}), we cannot quantify this finding and leave a more detailed analysis to future work.

With respect to quality, our results confirm previous findings that post-editing leads to similar or better translations \citep[\eg][]{Green2013}. An interesting nuance is that we find a slight, but consistent decrease in textual coherence within post-edited translations in both language pairs. As the research community is increasingly focussing on document-level MT, translation workbench providers will need to ensure integrability for future experimentation in real-life settings.

\section{Conclusion}
\label{sec:Conclusion}

We have assessed the impact of NMT on translation speed and quality in the banking and finance domain. Despite working with language pairs that have received limited attention in research contexts and employing a simple means of domain adaptation, the use of NMT enabled professional translators to work faster: 59.74\percent in DE--FR and 9.26\percent in DE--IT. Unlike a number of previous studies, these improvements are not relative to translation from scratch, but to translation with domain-specific TMs and TBs within a customary translation workbench, which sets a higher baseline in terms of translation speed.

NMT did not have a negative impact on quality. To the contrary, scores assigned by expert raters were slightly higher for post-edited DE--IT translations. Screen recordings showed that IT translators devoted more time to stylistic changes of NMT output, underpinning the importance of translator training in cases where NMT is to be used to optimise throughput rather than quality.

Another factor that likely contributed to the difference between time savings in DE--FR and DE--IT is that roughly half as much in-domain training data was available for the latter. While further investigation will be needed to determine the impact of in-domain data volume and more advanced domain adaptation techniques, our results suggest that NMT has the potential of increasing translation productivity even with complex text types, little-researched language pairs, and limited amounts of in-domain training data. The present study contributes empirical evidence for DE--FR and DE--IT translation of financial texts, and we hope to encourage similar investigations with other languages and domains.

\bibliographystyle{acl}
\fontsize{10.1pt}{10.1pt}\selectfont
\bibliography{refs}




\end{document}